\documentclass{applemlr}

\usepackage{amsmath}
\usepackage{enumerate}
\usepackage{algorithm}
\usepackage{algpseudocode}
\usepackage{amsfonts}
\usepackage{amsthm}
\usepackage{diagbox}
\usepackage{colortbl}
\usepackage{amssymb}
\usepackage{xspace}
\usepackage{wrapfig}
\usepackage{adjustbox}
\usepackage{tabularx}
\usepackage{booktabs}
\usepackage{mathtools}
\usepackage{tikz}
\usepackage{enumitem}
\usepackage{silence}
\usepackage{dsfont}
\usepackage[table]{xcolor}
\usepackage[dvipsnames]{xcolor}
\usepackage{multirow}
\usepackage{makecell}
\usepackage{xfakebold}

\usepackage{amsmath,amsfonts,bm}

\def\eqref#1{equation~\ref{#1}}

\def\1{\bm{1}}

\def\vc{{\bm{c}}}

\def\vo{{\bm{o}}}

\def\vu{{\bm{u}}}

\def\vx{{\bm{x}}}
\def\vy{{\bm{y}}}
\def\vz{{\bm{z}}}

\def\mI{{\bm{I}}}

\def\mW{{\bm{W}}}

\DeclareMathAlphabet{\mathsfit}{\encodingdefault}{\sfdefault}{m}{sl}
\SetMathAlphabet{\mathsfit}{bold}{\encodingdefault}{\sfdefault}{bx}{n}

\newcommand{\E}{\mathbb{E}}

\newcommand{\R}{\mathbb{R}}

\definecolor{textgray}{HTML}{6E6E73}
\usetikzlibrary{positioning, calc}
\usetikzlibrary{decorations.pathmorphing}

\makeatletter
\patchcmd{\wrong@fontshape}{\@gobbletwo}{}{}{}
\makeatother
\numberwithin{equation}{section}
\makeatletter
\AtBeginDocument{
  \urlstyle{sf}
  
}
\makeatother

\definecolor{light}{RGB}{125, 125, 125}
\crefname{tcb@cnt@pbox}{code}{code}
\Crefname{tcb@cnt@pbox}{Code}{Code}
\crefname{assumption}{assumption}{assumption}
\Crefname{assumption}{Assumption}{Assumptions}

\newtcolorbox[auto counter]{pbox}[2][]{
  colback=white,
  title=Code~\thetcbcounter: #2,
  #1,fonttitle=\sffamily,
  fontupper=\sffamily,
  arc=2pt,
  colframe=bgcolor,
  coltitle=fgcolor,
  colbacktitle=bgcolor,
  toptitle=0.25cm,
  bottomtitle=0.125cm
}

\makeatletter
\newcommand\applefootnote[1]{%
  \begingroup
  \renewcommand\thefootnote{}%
  \renewcommand\@makefntext[1]{\noindent##1}%
  \footnote{#1}%
  \addtocounter{footnote}{-1}%
  \endgroup
}
\makeatother

\definecolor{cverbbg}{gray}{0.90}

\usepackage{url}
\usepackage{nicefrac}
\usepackage{microtype}

\newcommand*\methodname{STARFlow2}

\newcommand*\pretzel{Pretzel}

\NewDocumentCommand{\ying}{ mO{} }{\textcolor{teal}{\textsuperscript{\textit{Ying}}\textsf{\textbf{\small[#1]}}}}

\usepackage{inconsolata}
\usepackage{listings}
\usepackage{array}
\usepackage{environ}

\crefformat{equation}{Eq.~(#2#1#3)}
\crefname{section}{§}{§§}
\Crefname{section}{§}{§§}

\title{STARFlow2: \\Bridging Language Models and Normalizing Flows for Unified Multimodal Generation}

\author[2]{Ying Shen}
\author[1]{Tianrong Chen}
\author[1]{Yuan Gao}
\author[1]{Yizhe Zhang}
\author[1]{Yuyang Wang}
\author[1]{Miguel Angel Bautista}
\author[1]{Shuangfei Zhai}
\author[1]{Josh Susskind}
\author[1]{Jiatao Gu}

\affiliation[1]{Apple}
\affiliation[2]{UIUC}

\abstract{
Unified multimodal models that understand, reason over, and generate interleaved text--image sequences remain structurally fragmented: existing approaches either sacrifice visual fidelity through discrete tokenization, impose structural asymmetry by combining causal text generation with iterative diffusion-based denoising, or degrade pretrained understanding when adapting vision-language models for generation.
    We observe that autoregressive normalizing flows are autoregressive Transformers---sharing the same causal mask, KV-cache mechanism, and left-to-right structure as LLMs---making them the most natural paradigm for truly unified multimodal generation that is continuous, single-pass, and purely causal.
    We present \methodname{}, built on the \pretzel{} architecture that vertically interleaves a frozen pretrained VLM stream with a TARFlow stream via residual skip connections, both operating under the same causal mask.
    This design simultaneously preserves pretrained multimodal understanding, enables high-fidelity continuous image generation, and achieves structural unification under a single causal mechanism.
    Combined with a deep-shallow flow design and a unified FAE latent space, \methodname{} supports cache-friendly interleaved generation where both text and visual outputs directly enter the KV-cache without re-encoding.
    Experiments demonstrate strong performance across image generation and multimodal understanding benchmarks, validating autoregressive flows as a viable foundation for unified multimodal modeling.
}

\metadata[Code]{\url{https://github.com/apple/ml-starflow}}
\metadata[Correspondence]{\sffamily \email{ying22@illinois.edu}; \email{jgu32@apple.com}}
\date{\sffamily\today}

\begin{document}

\maketitle

\begin{figure}[h!]
\centering
\includegraphics[width=\textwidth]{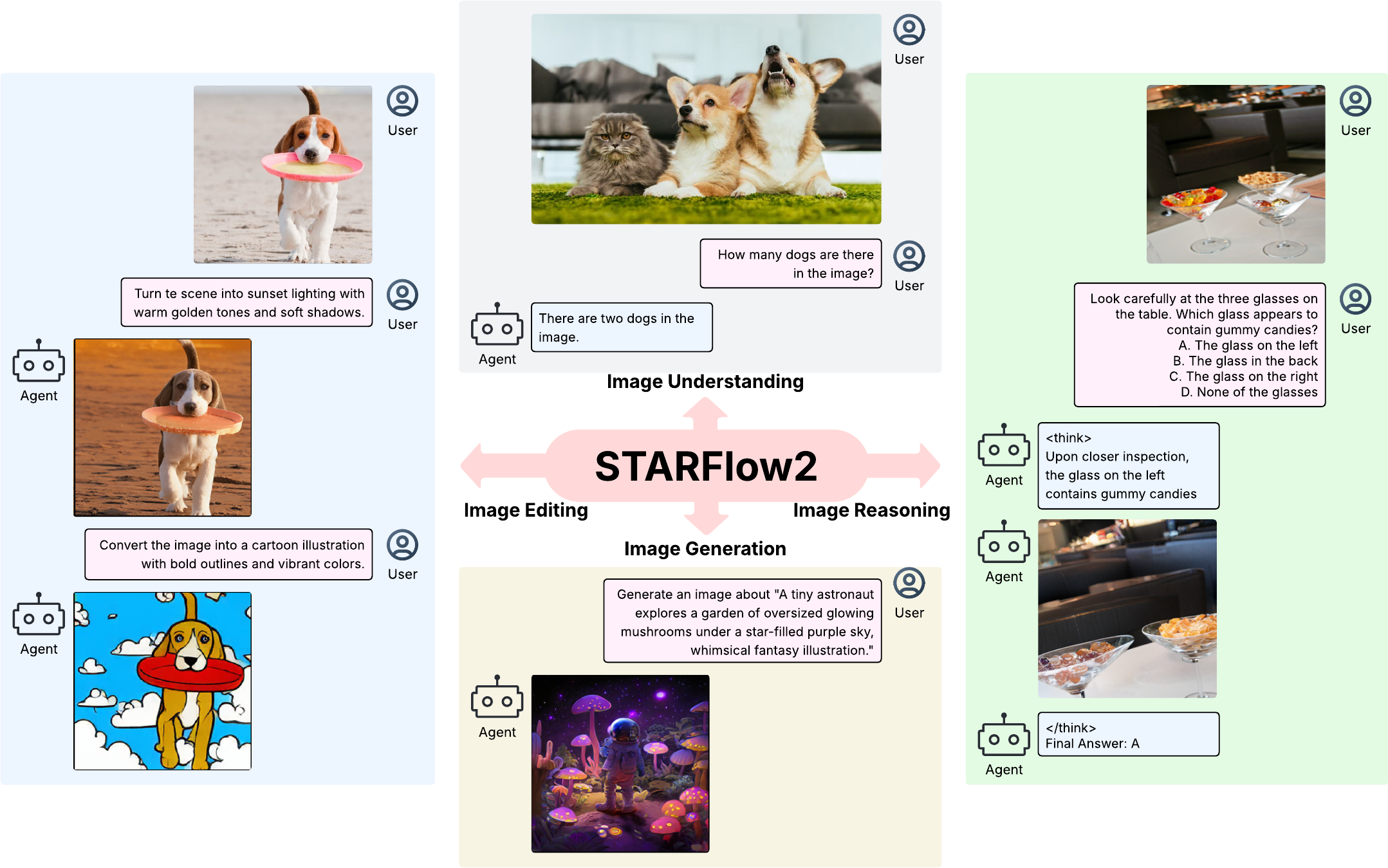}
\caption{
\textbf{\methodname{} as a unified multimodal architecture.}
A single model supports image generation, editing, understanding, and reasoning across diverse image-centric tasks.
}
\label{fig:teaser}
\end{figure}

\section{Introduction}

Unified multimodal models that perceive, reason over, and generate interleaved text--image sequences have emerged as a key goal toward general-purpose AI~\citep{zhou2024transfusion,wang2024emu3,deng2025emerging,xie2025show}.
By treating images and text as interleaved steps in a shared generation sequence, such models can support interactive multi-turn editing~\citep{ge2024seed,zhou2025multi} and problem solving with visual thoughts~\citep{hu2024visual,chern2025thinking}.

Despite growing interest, existing ``unified'' multimodal models are not truly unified in their generation mechanisms.
One line of work discretizes images into tokens and trains a single language model over the joint text-image sequence~\citep{wang2024emu3,li2025onecat,chen2025janus,chen2025blip3}. While architecturally elegant, this approach sacrifices the continuous nature of visual data---quantization introduces information loss and limits generation fidelity~\citep{luo2024open,wang2025bridging}.
A more popular paradigm combines autoregressive language modeling for text with diffusion-based denoising for images within a single backbone~\citep{zhou2024transfusion,xie2024show,xie2025show,shi2024lmfusion,liu2025tuna,deng2025emerging}.
However, these two generation mechanisms are structurally different: text tokens are generated causally under a left-to-right mask, while images require iterative denoising often with different attention patterns.
Generated images cannot directly enter the causal KV-cache as reusable context---a separate re-encoding step is needed for interleaved generation.
Mixture-of-Transformers (MoT)~\citep{liang2024mixture}, adopted in BAGEL~\citep{deng2025emerging}, routes different modalities to modality-specific feed-forward parameters while sharing attention.
Though this appears unified, it remains two specialized sub-networks sharing only attention within a single Transformer backbone.
Moreover, as we show empirically (\cref{sec:mot_comparison}), MoT faces an inherent dilemma when combined with TARFlow: freezing the VLM leads to poor generation quality, while finetuning the VLM degrades multimodal understanding.

We argue that a truly unified architecture must simultaneously satisfy three desiderata:
\begin{enumerate}[leftmargin=*,nosep,label=\textbf{(D\arabic*)}]
    \item \textbf{Preserve pretrained VLM understanding}---retain the strong multimodal perception and reasoning capabilities of a pretrained vision-language model without degradation from generation training.
    \item \textbf{High-fidelity continuous image generation}---generate images in continuous latent space without quantization loss, maintaining visual quality comparable to dedicated generative models.
    \item \textbf{Structurally unified causal generation}---generate both text and images under the same causal mechanism (same mask, same KV-cache, single-pass decoding), without diffusion's iterative denoising or re-encoding overhead.
\end{enumerate}
Discrete tokenization violates (D2); diffusion hybrids violate (D3); and MoT, depending on training strategy, violates either (D1) or (D2).

Recently, STARFlows~\citep{zhainormalizing,gu2025starflow,gu2025starflowv} have shown that normalizing flows, when parameterized by causal Transformers, can generate continuous visual data with quality matching or exceeding diffusion models.
Crucially, these models generate token-by-token from left to right---using the \textit{same} causal mask, the \textit{same} KV-cache mechanism, and the \textit{same} autoregressive structure as LLMs.
The only difference is the output head: instead of predicting discrete token logits, the flow predicts affine transformation parameters for continuous latents.
In other words, there is \textit{no structural gap} between \textit{autoregressive flows} and \textit{language models}---making flows a natural paradigm to satisfy (D2) and (D3) simultaneously: continuous, single-pass, and purely causal.

\begin{figure*}[t]
    \centering
    \includegraphics[width=\linewidth]{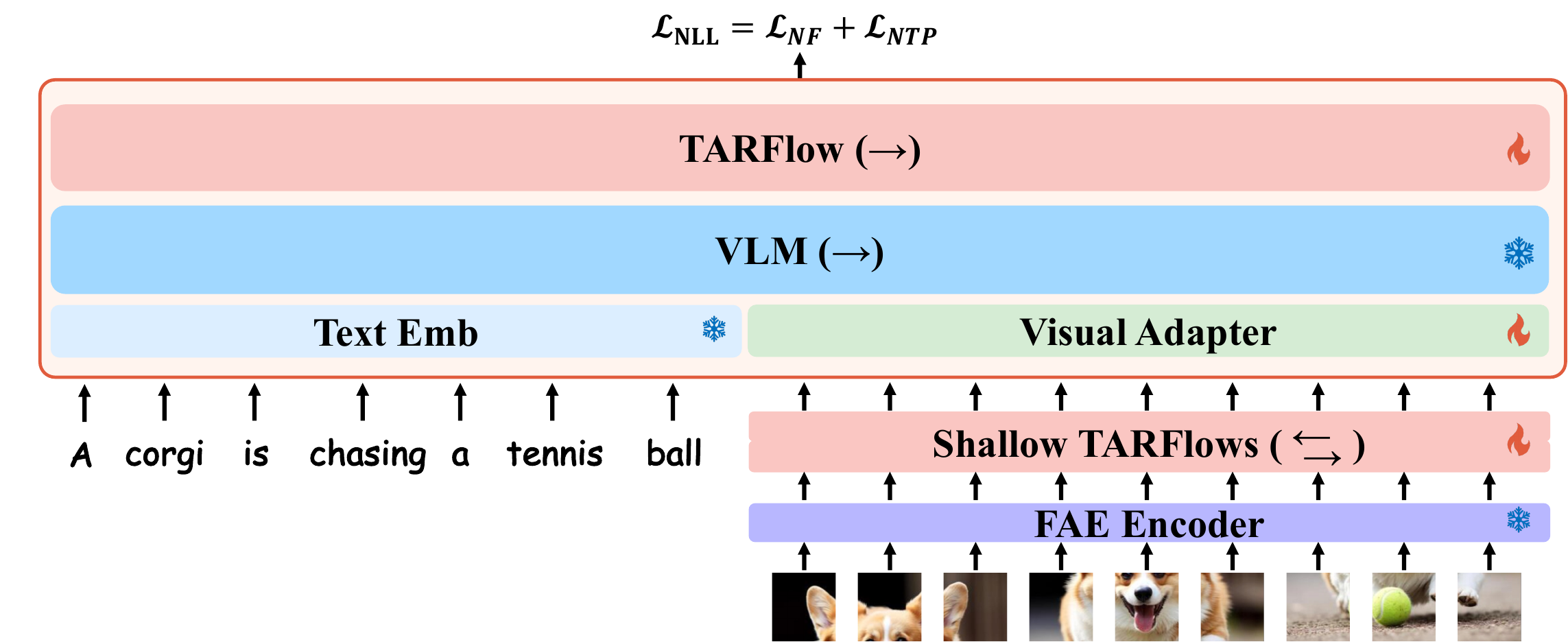}
    \vspace{-4mm}
    \caption{\textbf{Overview of the \pretzel{} architecture in \methodname{}.} A VLM stream and a TARFlow stream are vertically interleaved via crossing skip connections, operating on the same multimodal sequence under a shared causal mask. Shallow TARFlows refine visual latents locally. The model is trained with a unified NLL objective.}
    \label{fig:overview}
    \vspace{-3mm}
\end{figure*}


Building on this insight, we introduce \methodname{}, a unified multimodal model built on the \textbf{\pretzel{} architecture}---named for the characteristic shape formed by its two streams crossing through vertical skip connections (\cref{fig:overview}).
\pretzel{} vertically interleaves a pretrained VLM stream (for language modeling and multimodal understanding) with a TARFlow stream (for continuous visual generation) via residual skip connections, satisfying (D1) by keeping the VLM frozen while enabling rich cross-modal interaction.
Both streams process the same interleaved multimodal sequence under the same causal mask, achieving true architectural unification (D3).
Unlike MoT's horizontal separation---where different tokens route to different parameters---\pretzel{} interleaves the two streams vertically, allowing both to attend over all tokens and exchange information through skip connections at every position.
Combined with a deep-shallow flow design~\citep{gu2025starflow} and a unified FAE latent space~\citep{gao2025one}, \methodname{} supports cache-friendly interleaved text-image generation without visual re-encoding, while maintaining the fidelity of continuous-space generation (D2) and exact likelihood training.

Our contributions are as follows:
\begin{itemize}[leftmargin=*,nosep]
    \item We present \methodname{}, the first unified multimodal framework where both text and image generation employ the same autoregressive Transformer mechanism under the same causal mask, enabling cache-friendly interleaved generation without quantization, iteration, or visual re-encoding~(D2, D3).
    \item We propose the \pretzel{} architecture that vertically interleaves a frozen pretrained VLM with a TARFlow backbone via residual skip connections---in contrast to MoT's horizontal modality separation---preserving pretrained understanding while enabling rich cross-modal interaction within a single causal sequence model~(D1).
    \item Experiments on multimodal understanding and image generation benchmarks demonstrate that \methodname{} simultaneously achieves strong performance across all three desiderata, validating autoregressive flows as a foundation for unified multimodal generation.
\end{itemize}

\section{Preliminaries}

\paragraph{Unified Multimodal Generation}

A unified multimodal model processes interleaved text--image sequences $\mathcal{C} = (\vc_1,\ldots,\vc_T)$, where each element $\vc_t$ is either a discrete text token or a continuous visual latent.
The goal is to support both multimodal understanding (image-conditioned text generation) and visual generation (text-conditioned image synthesis) within a single model.
Most current approaches build on pretrained vision-language models (VLMs) that already achieve strong multimodal understanding~\citep{liu2024improved,Qwen25-VL}, and augment them with image generation capabilities.
The central challenge is how to integrate visual generation without degrading the VLM's pretrained understanding or introducing structural asymmetry between modalities.

\paragraph{Feature Auto-Encoder (FAE)}

\methodname{} operates in the latent space of a Feature Auto-Encoder (FAE)~\citep{gao2025one}, which provides a compact continuous representation serving both understanding and generation.
We train FAE on DINOv2-g/14~\citep{oquab2023dinov2} features, which we find better suited for generation than SIGLIP-based representations while retaining strong understanding performance.
Given an image, the FAE encoder produces visual latents $\vx \in \mathbb{R}^{N \times D}$, where $N$ is the number of visual tokens and $D$ is the latent dimensionality.
This shared latent space enables a single representation to serve as both the conditioning input for multimodal understanding and the generation target for normalizing flows.

\paragraph{Autoregressive Normalizing Flows}

Normalizing flows (NFs)~\citep{dinh2014nice,rezende2015variational,dinh2016density,kingma2018glow,ho2019flow++} are likelihood-based generative models that learn an invertible mapping between a simple distribution (e.g., a standard Gaussian) and a complex data distribution. In particular, given a continuous input $\vx \sim p_{\textrm{data}}, \vx \in \R^D$, an NF learns a bijection $f_\theta: \R^D \rightarrow \R^D$ that maps data $\vx$ to latents $\vz = f_\theta(\vx)$. Derived from the change-of-variables formula, NFs can be trained end-to-end via a tractable maximum-likelihood objective:
\begin{equation}
    \mathcal{L}_\textrm{NF}(\theta) = -\E_{\vx } \left[
    \log p_0(f_\theta(\vx)) + \log|\textrm{det}(J_{f_\theta}(\vx))|
    \right],
\end{equation}
where the first term encourages mapping data to high-density regions of a simple prior $p_0$, and the Jacobian term $J_f$ accounts for the local volume change induced by $f_\theta$, preventing the model from collapsing. 
Once trained, one automatically obtains a generative model by inverting $f_\theta$, with a sampling process: $\vz \sim p_0(\vz), \vx = f^{-1}_\theta(\vz)$.

Recently, TARFlow-style models~\citep{zhainormalizing,gu2025starflow,gu2025starflowv} have revived normalizing flows for generative modeling by parameterizing them with causal Transformers.
Specifically, they instantiate Autoregressive Flows (AFs)~\citep{kingma2016improved,papamakarios2017masked} by stacking multiple invertible autoregressive flow (AF) blocks with alternating orderings. 
Given an input presented in the form of a sequence $\vx \in \R^{N\times D}$, where $N$ is the sequence length and $D$ is the dimension, each AF block applies an affine transform whose parameters are predicted by a causal Transformer under a self-exclusive causal mask for both forward ($\vx \rightarrow \vz$) and sampling ($\vz \rightarrow \vx$) process:
\begin{align}
\label{eq:affine}
    \vz_n = \left(\vx_n - \mu_\theta(\vx_{< n})\right) / \sigma_\theta(\vx_{< n}), \quad
    \vx_n = \mu_\theta(\vx_{< n}) + \sigma_\theta(\vx_{< n}) \cdot \vz_n,
\end{align}
where $\vx, \vz$ are the input and output of each block. This can be viewed as "next-token prediction" with affine transformation.
STARFlow~\citep{gu2025starflow} introduces a deep-shallow architecture, where a deep AF block captures most of the model's capacity, followed by a few shallow AF blocks that further refine the image generation.
Note that if we have the deep AF block to follow the left-to-right causal order, it inherits the same causal structure as language models, making them a natural candidate for unifying continuous visual generation with discrete text modeling in an autoregressive manner.
\section{\methodname{}}

This section details the three components of \methodname{}: the \pretzel{} architecture that vertically interleaves a pretrained VLM with a TARFlow stream (\cref{sec:pretzel}); the deep-shallow flow design that factorizes visual generation into global multimodal modeling and local refinement (\cref{sec:deep_shallow}); and the multi-stage training pipeline that progressively activates components (\cref{sec:pipeline}).

\subsection{The \pretzel{} Architecture}
\label{sec:pretzel}

The core of \methodname{} is the \pretzel{} architecture, which vertically interleaves two autoregressive streams---a pretrained VLM and a TARFlow stream---connected by residual skip connections.
Both streams process the same interleaved multimodal sequence $\mathcal{C} = (\vc_1,\ldots,\vc_T)$ under a single left-to-right causal mask, where each element $\vc_t$ is either a text token or a visual latent.

\paragraph{VLM Stream.}
The VLM stream is initialized from a pretrained vision-language model (Qwen2.5-VL-7B) and provides high-level semantic representations for language modeling and multimodal understanding.
For text positions $t \in \mathcal{M}$, the token is mapped to an embedding via the pretrained text embedding layer.
For visual positions $t \in \mathcal{N}$, the intermediate visual latents $\vu$ (produced by the shallow flow blocks, described in \cref{sec:deep_shallow}) are projected by a lightweight adapter into the VLM representation space.
The VLM processes the full interleaved sequence and produces contextual hidden states $\vy_{\mathrm{vlm}}$.

\paragraph{TARFlow Stream.}
The TARFlow stream is an autoregressive flow block that operates on the same multimodal sequence under the same causal mask.
For each visual latent $\vu_t$, where $t \in \mathcal{N}$, it applies the autoregressive affine transformation defined in \cref{eq:affine}, predicting location and scale parameters conditioned on all preceding tokens in the multimodal sequence.
For text positions, the TARFlow stream performs standard causal sequence modeling.
Because both the VLM and TARFlow streams use the same left-to-right causal structure, they are architecturally compatible---this is what enables true unification.
\begin{figure*}[!t]
    \centering
    \includegraphics[width=\linewidth]{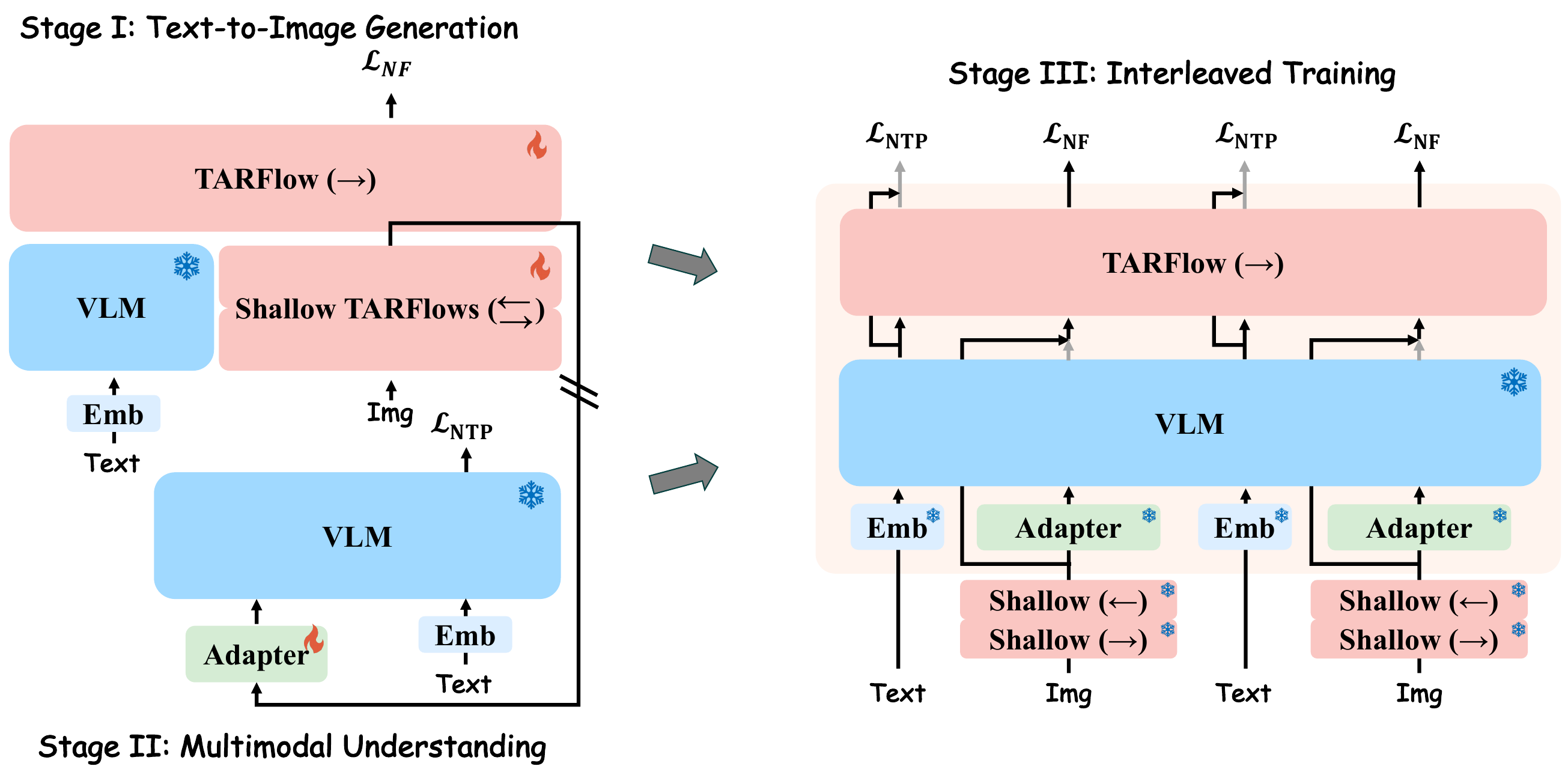}
    \vspace{-2mm}
    \caption{\textbf{Multi-Stage Training Pipeline of \methodname{}.}
    \textit{Stage 1}: Train the TARFlow stream and shallow blocks on text-image pairs for text-to-image generation (VLM frozen).
    \textit{Stage 2}: Align the visual representation with the VLM by training the adapter on image-to-text tasks (shallow blocks and VLM frozen).
    \textit{Stage 3}: Activate the vertical skip connections of the \pretzel{} architecture and jointly optimize on a mixture of understanding, generation, editing, and interleaved tasks.
    }
    \vspace{-3mm}
    \label{fig:pipeline}
\end{figure*}
\paragraph{Vertical Skip Connections.}
The two streams exchange information through skip connections at every position---the defining feature of the \pretzel{} architecture (see Stage 3 in \cref{fig:pipeline}).
Specifically, the TARFlow stream input and output head are defined per-position as:
\begin{align}
\label{eq:vis_res}
    \text{TARFlow input:} \quad \hat{\vc}_t &= \begin{cases} \vu_t + \mW_{\mathrm{vlm}} \cdot \vy_{\mathrm{vlm},t} & \text{if } t \in \mathcal{N} \text{ (visual)} \\ \vy_{\mathrm{vlm},t} & \text{if } t \in \mathcal{M} \text{ (text)} \end{cases} \\[4pt]
\label{eq:txt_res}
    \text{Output sample:} \quad \hat{\vo}_t &= \begin{cases} \mathcal{N}\big( \mu_\mathcal{D}(\vy_{\mathcal{D},t}),\, \sigma^2_\mathcal{D}(\vy_{\mathcal{D},t})\big) & \text{if } t \in \mathcal{N} \text{ (visual)} \\ \mathrm{Cat}\big( \mathrm{softmax} \big( \mathrm{LM}\left( \vy_{\mathrm{vlm},t} + \mW_{\mathcal{D}} \cdot \vy_{\mathcal{D},t}\right)\big)\big) & \text{if } t \in \mathcal{M} \text{ (text)} \end{cases}
\end{align}
where $\mW_{\mathrm{vlm}}$ and $\mW_{\mathcal{D}}$ are zero-initialized linear projections, $\vy_{\mathrm{vlm},t}$ and $\vy_{\mathcal{D},t}$ denote the VLM and TARFlow stream output at position $t$.
The visual skip connection at the TARFlow input preserves the low-level visual information in $\vu_t$ while injecting high-level semantic information from the VLM into the TARFlow stream. 
For visual position at the output, the last-layer Deep TARFlow hidden state is projected to predict the affine parameters $(\mu_{\mathcal{D}}, \sigma_{\mathcal{D}})$ to induce the Gaussian distribution of $\mathcal{N}( \mu_\mathcal{D},\, \sigma^2_\mathcal{D})$ over the intermediate visual latent. 
For text position, the language modeling head $\mathrm{LM}(\cdot)$ maps the fused text representation to vocabulary logits, which define a categorical distribution ($\mathrm{Cat}(\cdot)$) over the next token.
The text skip connection preserves the pretrained language modeling behavior of the VLM while allowing the Deep TARFlow to learn multimodal corrections.
Both projections are zero-initialized so that \methodname{} starts from the pretrained VLM and flow behaviors, gradually learning cross-modal corrections during training.

\subsection{Deep-Shallow Flow Design}
\label{sec:deep_shallow}

A single autoregressive pass cannot fully capture the distribution of FAE latents, which exhibit strong local spatial correlations that a purely left-to-right model would need excessive depth to absorb.
Following STARFlow~\citep{gu2025starflow}, \methodname{} addresses this with a deep-shallow flow design that factorizes the generative process into two stages.
A stack of visual-only shallow AF blocks ($f_\mathcal{S}$) with alternating scan directions first transforms FAE latents into simpler intermediate representations $\vu = f_\mathcal{S}(\vx)$ that can be effectively modeled by a single autoregressive pass.
The TARFlow stream ($f_\mathcal{D}$), within the \pretzel{} architecture, then models $\vu$ conditioned on the full multimodal context.
This factorization is essential: as shown in~\citet{gu2025starflow}, the shallow blocks absorb the local complexity of the visual distribution, enabling the deep block to focus on global structure and cross-modal dependencies.

The composed flow yields an exact log-likelihood objective:
\begin{equation}
\label{eq:img_nll}
    p(\vx)
    = p_0(\vz) \left|\det J_{f_\mathcal{D}}(\vu; \mathcal{C})\right| \left|\det J_{f_\mathcal{S}}(\vx)\right|,
\end{equation}
where $\vz = f_{\mathcal{D}}(\vu; \mathcal{C})$ and $p_0$ is a standard Gaussian prior. Both the shallow blocks and TARFlow stream contribute to the likelihood computation.
Crucially, the shallow blocks operate exclusively on visual latents and do not interfere with the left-to-right causal structure of the \pretzel{} architecture, preserving cache-friendly interleaved generation.

\subsection{Multi-Stage Training Pipeline}
\label{sec:pipeline}

We adopt a multi-stage training paradigm that progressively activates components of \pretzel{}.

\paragraph{Stage 1: Text-to-Image Generation.}
We first establish a strong visual generation backbone by training on large-scale text-image pairs for text-to-image generation.
We optimize the TARFlow stream $f_{\mathcal{D}}$ and the shallow blocks $f_{\mathcal{S}}$, while keeping the pretrained VLM frozen.
The VLM encodes text captions into contextual representations that condition the flow, but receives no gradient updates.
The training objective minimizes the negative log-likelihood of the composed flow:
\begin{align}
\label{eq:img_loss}
    \mathcal{L}_{\mathrm{NF}}
    &= \mathbb{E}_{\vx}\left[\sum_{n=1}^{N} \left(\frac{1}{2}\|\vz_n\|^2 + \log \sigma_\mathcal{D}(\vu_{<n}; \vc)\right) - \log\left|\det J_{f_\mathcal{S}}(\vx)\right|\right] \nonumber \\
    &= \mathbb{E}_{\vx}\left[\sum_{n=1}^{N} -\log \mathcal{N}(\vu_n;\, \mu_\mathcal{D}(\vu_{<n}; \vc),\, \sigma_\mathcal{D}^2(\vu_{<n}; \vc)) - \log\left|\det J_{f_\mathcal{S}}(\vx)\right|\right],
\end{align}
where $\vu = f_\mathcal{S}(\vx)$, $\vz_n = (\vu_n - \mu_\mathcal{D}) / \sigma_\mathcal{D}$, and $\vc$ denotes the preceding multimodal context (e.g., the text caption in Stage 1).
The second line reveals that the TARFlow stream performs \textbf{Next Gaussian Prediction (NGP)} in $\vu$-space---the continuous counterpart of next-token prediction: at each visual position, the model predicts the mean and scale of a Gaussian over the next latent $\vu_n$, conditioned on all preceding tokens, just as an LLM predicts a categorical distribution over the next text token.
At inference, sampling from this predicted Gaussian yields:
\begin{equation}
\label{eq:sampling}
    \vu_n = \mu_\mathcal{D}(\vu_{< n}; \vc) + \sigma_\mathcal{D}(\vu_{< n}; \vc) \cdot \vz_n, \quad \vz_n \sim \mathcal{N}(0, \mI).
\end{equation}

\paragraph{Stage 2: Multimodal Understanding.}
With the flow components trained, we align the intermediate visual representation $\vu$ with the pretrained VLM so that it can serve as visual input for multimodal understanding.
We train on image-to-text data including captioning and multimodal understanding tasks.
We freeze the shallow blocks and VLM, and optimize only the adapter that maps $\vu$ into the VLM representation space using the next-token prediction loss:
\begin{equation}
\label{eq:ntp_stage2}
    \mathcal{L}_{\mathrm{NTP}}
    =
    -
    \frac{1}{|\mathcal{M}|}
    \sum_{t \in \mathcal{M}}
    \log p_\theta
    \left(
        y_t \mid \mathcal{C}_{<t}
    \right).
\end{equation}
Optionally, we can also distill from the frozen VLM (with its original visual encoder) to further improve alignment.
This stage ensures the FAE latent space, originally designed for generation, also supports understanding through the VLM.

\paragraph{Stage 3: Interleaved Generation and Understanding.}
In the final stage, we activate the vertical skip connections of the \pretzel{} architecture and jointly train on a mixture of data covering multimodal understanding, text-to-image generation, image editing, and interleaved text-image generation.
Since both projections $\mW_{\mathrm{vlm}}$ and $\mW_{\mathcal{D}}$ are zero-initialized, \methodname{} starts from the pretrained behaviors of Stages 1--2 and gradually learns cross-modal corrections.
The joint objective combines the flow loss and next-token prediction:
\begin{equation}
\label{eq:ce_text_loss}
    \mathcal{L} = \mathcal{L}_{\mathrm{NF}} + \lambda\,\mathcal{L}_{\mathrm{NTP}},
\end{equation}
where $\lambda$ balances the two modality losses.
This stage unifies all capabilities---understanding, generation, editing, and interleaved synthesis---within the \pretzel{} framework, with all components jointly optimized end-to-end.

\section{Experimental Setup}

 \paragraph{Datasets}
We construct a collection of text-image datasets to support the multi-stage training of \methodname{}. In Stage 1, we focus on establishing a strong text-to-image generation backbone using large-scale image-caption data, including an in-house dataset along with CC12M~\citep{changpinyo2021conceptual}, and JourneyDB~\citep{sun2023journeydb}, totaling around 800M text--image pairs.
In Stage 2, we train the visual adapter for multimodal understanding using a mixture of CC12M and Cambrian-7M~\citep{tong2024cambrian}, an instruction-style visual question answering data. This stage is trained on approximately 200M examples for image-to-text generation.
In Stage 3, we further train \methodname{} on a broader mixture of datasets covering multimodal understanding, image generation, editing, and interleaved text-image generation datasets, including the in-house dataset in Stage 1, BLIP3-o-60K~\citep{chen2025blip3}, Cambrian-7M~\citep{tong2024cambrian}, CoMM~\citep{chen2025comm}, Pico-Banana~\citep{qian2025pico}, OmniEdit~\citep{wei2024omniedit}, and Zebra-CoT~\citep{li2025zebra}. This final stage is trained on approximately 80M examples.
\vspace{-1em}


\paragraph{Evaluation}
We evaluate \methodname{} on several multimodal understanding benchmarks: MME~\citep{fu2025mme}, SEED-Bench~\citep{li2023seed}, MMBench~\citep{liu2024mmbench}, MMMU~\citep{yue2024mmmu}
to assess general multimodal perception and reasoning capability, and GQA~\citep{hudson2019gqa} for real-world visual reasoning and AI2D~\citep{kembhavi2016diagram} for scientific diagram comprehension.
For visual generation, we evaluate our model on two widely used benchmarks: GenEval~\citep{ghosh2023geneval} and DPG-Bench~\citep{hu2024ella}.

\vspace{-1em}


\paragraph{Model and Training Details}

We employ Qwen2.5-VL-7B-Instruct~\citep{Qwen25-VL} as the pretrained VLM and FAE~\citep{gao2025one} trained on DINOv2-g/14~\citep{oquab2023dinov2} features as the image encoder. The pretrained VLM and the FAE encoder are kept frozen throughout all training stages.
We follow the STARFlow~\citep{gu2025starflow} design for the causal Deep TARFlow stream and the two visual-only shallow TARFlow blocks. 
To align flow-based visual latents with the VLM representation space, we introduce a FiLM-style~\citep{perez2018film} adapter, which first projects visual latents through a lightweight MLP stack and then applies adaptive LayerNorm modulation conditioned on the noise level. 
In addition, we adopt the multi-noise training strategy from iTARFlow~\citep{chen2026normalizing} for visual generation.
These altogether result in a total of 3.6B trainable parameters.
All models are trained at 256 × 256 resolution with a global batch size of 1024. More details can be found in \cref{sec:imp}.

\section{Results}

\subsection{Quantitative Results}

\begin{table*}[t]
\centering
\small
\setlength{\tabcolsep}{5pt}
\resizebox{\linewidth}{!}{%
\begin{tabular}{llc cccccc}
\toprule
Types & Models & \# Params. & MME-p$\uparrow$ & GQA$\uparrow$ & SEED$\uparrow$ & MMB(en)$\uparrow$ & MMMU(val)$\uparrow$ & AI2D$\uparrow$ \\
\midrule

\multirow{3}{*}{Und. Only}
& LLaVA-v1.5~\citep{liu2024improved} & 7B & 1510.7 & 62.0 & 58.6 & 64.3 & -- & -- \\
& Qwen-VL-Chat~\citep{qwenvl} & 7B & 1487.6 & 57.5 & 58.2 & 60.6 & -- & 57.7 \\
& Qwen-2.5-VL-Instruct~\citep{Qwen25-VL} & 7B & 1677.9 & 60.7 & 75.5 & 83.8 & 50.6 & 82.3 \\

\midrule
\multirow{3}{*}{\begin{tabular}[c]{@{}l@{}}Composite\\Unified\end{tabular}}
& ILLUME~\citep{wang2025illume} & 7B & 1445.3 & -- & 72.9 & 75.1 & 38.2 & 71.4 \\
& BLIP3-o~\citep{chen2025blip3} & 8B & 1682.6 & -- & 77.5 & 75.5 & 50.6 & -- \\
& SEED-X~\citep{ge2024seed} & 17B & 1457.0 & 49.1 & 66.5 & 70.1 & 35.6 & -- \\

\midrule
\multirow{8}{*}{Native Unified}
& TUNA~\citep{liu2025tuna} & 1.5B & 1461.5 & 61.4 & 69.3 & -- & 39.1 & 71.4 \\
& Janus-Pro~\citep{chen2025janus} & 7B & 1567.1 & 62.0 & 72.1 & 79.2 & 41.0 & -- \\
& Mogao~\citep{liao2025mogao} & 7B & 1592.0 & 60.9 & 74.6 & 75.0 & 44.2 & -- \\
& Show-o2~\citep{xie2025show}  & 7B & 1620.5 & 63.1 & 69.8 & 79.3 & 48.9 & 78.6 \\
& TUNA~\citep{liu2025tuna} & 7B & 1641.5 & 63.9 & 74.7 & -- & 49.8 & 79.3 \\
& Emu3~\citep{wang2024emu3} & 8B & -- & 60.3 & 68.2 & 58.5 & 31.6 & 70.0 \\
& BAGEL~\citep{deng2025emerging} & 14B & 1687.0 & -- & -- & 85.0 & 55.3 & -- \\
\rowcolor{blue!10} \cellcolor{white}& \methodname{} (Ours) & 10.6B & 1528.8 & 55.8 & 71.1 & 71.5 & 44.7 & 67.7 \\

\bottomrule
\end{tabular}}
\caption{\textbf{Evaluation on multimodal understanding benchmarks.}}
\vspace{-2mm}
\label{tab:und}
\end{table*}
\begin{table*}[t]
\centering
\small
\setlength{\tabcolsep}{5pt}
\resizebox{\linewidth}{!}{%
\begin{tabular}{llc ccccccc c}
\toprule
Type & Method & \# Params. &
Single Obj. & Two Obj. & Counting & Colors & Position & Color Attri. & Overall$\uparrow$ \\
\midrule

\multirow{3}{*}{Gen. Only}
& SD3-Medium~\citep{esser2024scaling} & 2B &
0.99 & 0.94 & 0.72 & 0.89 & 0.33 & 0.60 & 0.74 \\
& FLUX.1 [dev]$^\dagger$~\citep{batifol2025flux} & 12B &
0.98 & 0.93 & 0.75 & 0.93 & 0.68 & 0.65 & 0.82 \\
& Qwen-Image~\citep{wu2025qwen} & 20B &
0.99 & 0.92 & 0.89 & 0.88 & 0.76 & 0.77 & 0.87 \\
\midrule

\multirow{4}{*}{\begin{tabular}[c]{@{}l@{}}Composite\\Unified\end{tabular}}
& MetaQuery-XL~\citep{pan2025transfer} & 7B &
-- & -- & -- & -- & -- & -- & 0.80 \\
& BLIP3-o~\citep{chen2025blip3} & 8B &
-- & -- & -- & -- & -- & -- & 0.84 \\
& UniWorld-V1$^\dagger$~\citep{lin2025uniworld} & 12B &
0.98 & 0.93 & 0.81 & 0.89 & 0.74 & 0.71 & 0.84 \\
& SEED-X~\citep{ge2024seed} & 17B &
0.97 & 0.58 & 0.26 & 0.80 & 0.19 & 0.14 & 0.49 \\

\midrule
\multirow{9}{*}{Native Unified}
& Transfusion~\citep{zhou2024transfusion} & 7B &
-- & -- & -- & -- & -- & -- & 0.63 \\
& Janus-Pro~\citep{chen2025janus} & 7B &
0.99 & 0.89 & 0.59 & 0.90 & 0.79 & 0.66 & 0.80 \\
& Mogao~\citep{liao2025mogao} & 7B &
1.00 & 0.97 & 0.83 & 0.93 & 0.84 & 0.80 & 0.89 \\


& Show-o2~\citep{xie2025show} & 7B &
1.00 & 0.87 & 0.58 & 0.92 & 0.52 & 0.62 & 0.76 \\
& TUNA~\citep{liu2025tuna} & 7B &
1.00 & 0.97 & 0.81 & 0.91 & 0.88 & 0.83 & 0.90 \\
& Emu3~\citep{wang2024emu3} & 8B &
-- & -- & -- & -- & -- & -- & 0.66 \\
& BAGEL~\citep{deng2025emerging} & 14B &
0.99 & 0.94 & 0.81 & 0.88 & 0.64 & 0.63 & 0.82 \\
& BAGEL$^\dagger$~\citep{deng2025emerging} & 14B &
0.98 & 0.95 & 0.84 & 0.95 & 0.78 & 0.77 & 0.88 \\
\rowcolor{blue!10} \cellcolor{white}
& \methodname{} (Ours) & 10.6B &
0.99 & 0.89 & 0.84 & 0.80 & 0.86 & 0.56 & 0.82 \\

\bottomrule
\end{tabular}}
\caption{\textbf{Evaluation of text-to-image generation on GenEval~\citep{ghosh2023geneval}.} $^\dagger$ refers to the method using LLM rewriters.}
\vspace{-4mm}
\label{tab:gen_eval}
\end{table*}
\begin{table*}[t]
\centering
\small
\setlength{\tabcolsep}{5pt}
\resizebox{\linewidth}{!}{%
\begin{tabular}{llc ccccc c}
\toprule
Type & Method & \# Params. &
Global & Entity & Attribute & Relation & Other & Overall$\uparrow$ \\
\midrule

\multirow{3}{*}{Gen. Only}
& SD3-Medium~\citep{esser2024scaling} & 2B &
87.90 & 91.01 & 88.83 & 80.70 & 88.68 & 84.08 \\
& FLUX.1 [dev]~\citep{batifol2025flux} & 12B &
82.10 & 89.50 & 88.70 & 91.10 & 89.40 & 84.00 \\
& Qwen-Image~\citep{wu2025qwen} & 20B &
91.32 & 91.56 & 92.02 & 94.31 & 92.73 & 88.32 \\
\midrule

\multirow{3}{*}{\begin{tabular}[c]{@{}l@{}}Composite\\Unified\end{tabular}}
& OmniGen2~\citep{wu2025omnigen2} & 7B &
88.81 & 88.83 & 90.18 & 89.37 & 90.27 & 83.57 \\
& BLIP3-o~\citep{chen2025blip3} & 8B &
-- & -- & -- & -- & -- & 81.60 \\
& UniWorld-V1~\citep{lin2025uniworld} & 12B &
83.64 & 88.39 & 88.44 & 89.27 & 87.22 & 81.38 \\

\midrule
\multirow{7}{*}{Native Unified}
& Janus-Pro~\citep{chen2025janus} & 7B &
86.90 & 88.90 & 89.40 & 89.32 & 89.48 & 84.19 \\
& Mogao~\citep{liao2025mogao} & 7B &
82.37 & 90.03 & 88.26 & 93.18 & 85.40 & 84.33 \\
& Show-o2~\citep{xie2025show} & 7B &
89.00 & 91.78 & 89.96 & 91.81 & 91.64 & 86.14 \\
& TUNA~\citep{liu2025tuna} & 7B &
90.42 & 91.68 & 90.94 & 91.87 & 90.73 & 86.76 \\
& Emu3~\citep{wang2024emu3} & 8B &
-- & -- & -- & -- & -- & 81.60 \\
& BAGEL~\citep{deng2025emerging} & 14B &
88.94 & 90.37 & 91.29 & 90.82 & 88.67 & 85.07 \\
\rowcolor{blue!10} \cellcolor{white}
& \methodname{} (Ours) & 10.6B &
91.45 & 91.83 & 88.91 & 91.09 & 88.61 & 84.94 \\

\bottomrule
\end{tabular}}
\caption{\textbf{Evaluation of text-to-image generation on DPG-Bench~\citep{hu2024ella}.}}
\vspace{-3mm}
\label{tab:dpg_bench}
\end{table*}

\paragraph{Multimodal understanding.}

We evaluate \methodname{} on multiple multimodal understanding benchmarks, as shown in \cref{tab:und}. \methodname{} achieves strong performance across standard benchmarks, including MME-P, GQA, SEED, MMBench, MMMU, and AI2D, demonstrating that the \pretzel{} architecture preserves the pretrained VLM's multimodal perception and reasoning capabilities (D1) while simultaneously supporting flow-based visual generation.
Note that \methodname{} is evaluated at $256 \times 256$ resolution due to the current FAE encoder constraint. Despite this limitation, the model maintains effective understanding performance, confirming that integrating a TARFlow stream through vertical skip connections does not compromise the frozen VLM's capabilities.


\paragraph{Image Generation.}

We further evaluate text-to-image generation on GenEval and DPG-Bench, as reported in \cref{tab:gen_eval,tab:dpg_bench}.
GenEval measures fine-grained instruction following across object presence, counting, colors, attributes, and spatial relationships, while DPG-Bench focuses on compositional text-to-image alignment at the global, entity, attribute, and relation levels.
\methodname{} achieves 0.82 on GenEval and 84.14 on DPG-Bench, demonstrating that autoregressive normalizing flows generate visually meaningful images (D2) while sharing the same causal decoding structure as text generation (D3). 

\paragraph{Effect of Joint Training on Interleaved Data}

We compare the text-to-image performance of \methodname{} after Stage~1 and Stage~3 on GenEval and DPG-Bench. Stage~1 trains the TARFlow stream and shallow TARFlows for text-to-image generation, while Stage~3 activates the vertical skip connections and jointly optimizes the model on multimodal understanding, image generation, editing, and interleaved generation data. 
As shown in \cref{tab:stage13_t2i}, Stage~3 improves image generation performance on both benchmarks, with relative gains of 60.8\% on GenEval and 3.6\% on DPG-Bench.
This indicates that joint multimodal training along with the vertical fusion in the \pretzel{} does not degrade the pretrained visual generation pathway.

\begin{figure*}[!t]
    \centering
    \includegraphics[width=\linewidth]{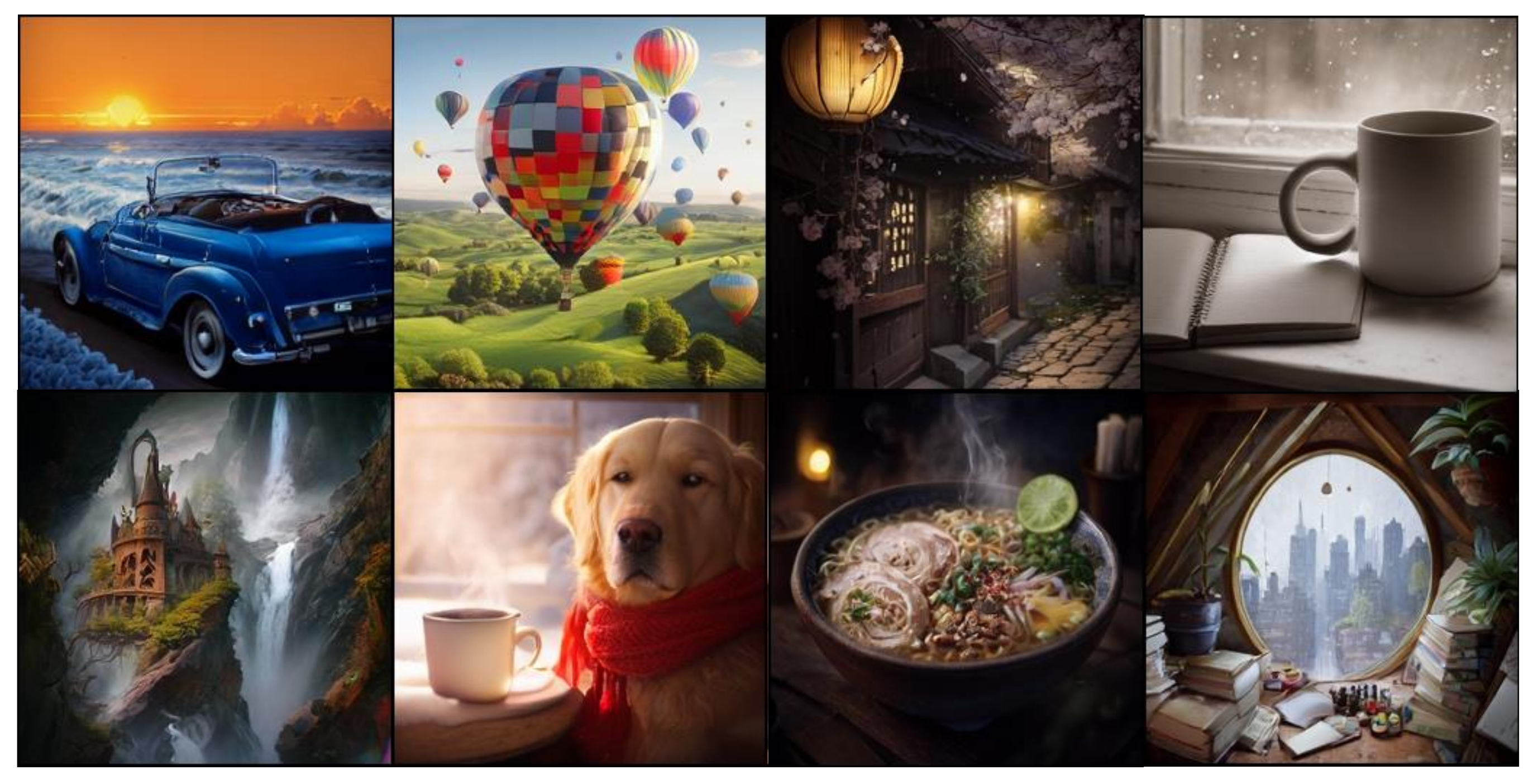}
    \vspace{-4mm}
    \caption{Text-to-Image generation examples from \methodname{} at 256 $\times$ 256 resolution.}
    \label{fig:t2i_ex}
    \vspace{-3mm}
\end{figure*}
\begin{figure*}[!t]
    \centering
    \includegraphics[width=\linewidth]{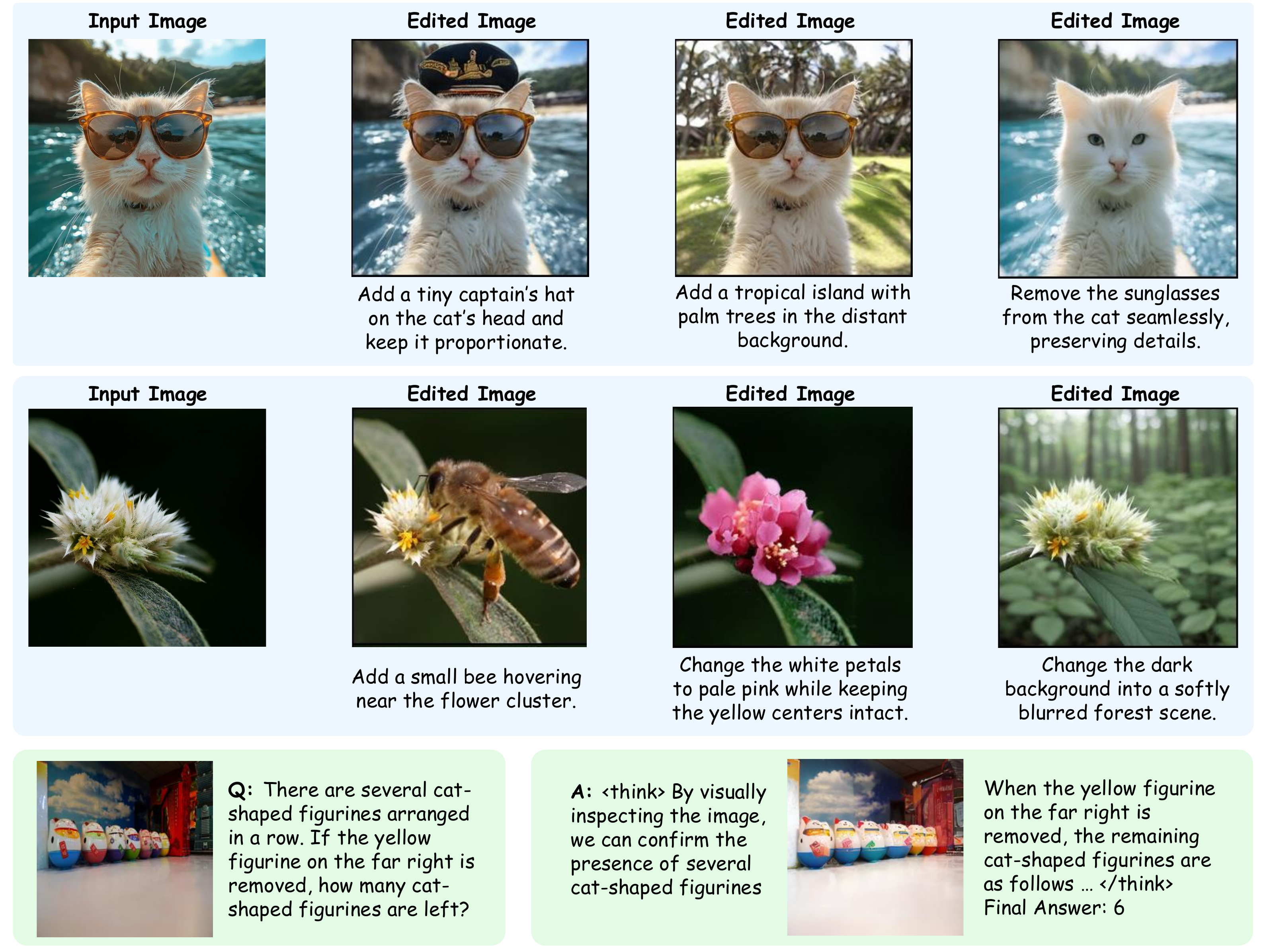}
    \vspace{-4mm}
    \caption{Image editing and interleaved text-image generation examples from \methodname{}.}
    \label{fig:interleave}
    \vspace{-3mm}
\end{figure*}

\subsection{Qualitative Results}
\begin{wraptable}{r}{0.52\linewidth}
\vspace{-1em}
\centering
\small
\setlength{\tabcolsep}{6pt}
\resizebox{\linewidth}{!}{%
\begin{tabular}{lcc}
\toprule
Training Stage & GenEval $\uparrow$ & DPG-Bench $\uparrow$ \\
\midrule
Stage 1: Text-to-Image Generation
& 0.51 & 82.02 \\
Stage 3: Interleaved Training
& \textbf{0.82 }& \textbf{84.94} \\
\midrule
$\Delta$ Improvement
& +0.31 & +2.92 \\
\bottomrule
\end{tabular}
}
\vspace{-0.5em}
\caption{
\textbf{Effect of interleaved training on text-to-image generation.}
We compare \methodname{} after Stage~1 and Stage~3 using the overall scores on GenEval~\citep{ghosh2023geneval} and DPG-Bench~\citep{hu2024ella}.
}
\label{tab:stage13_t2i}
\vspace{-4mm}
\end{wraptable}

\cref{fig:t2i_ex} shows representative text-to-image generation examples from \methodname{}.
\cref{fig:interleave} demonstrates examples in image editing and interleaved text-image generation.
The qualitative results show that \methodname{} can follow editing instructions, modify local attributes, and adjust visual content while preserving the overall scene structure, verifying it as a unified multimodal generator with cache-friendly interleaved generation.

\subsection{Pretzel vs.\ Bagel (MoT)}
\label{sec:mot_comparison}

\begin{wrapfigure}{r}{0.4\linewidth}
    \centering
    \vspace{-4mm}
    \includegraphics[width=\linewidth]{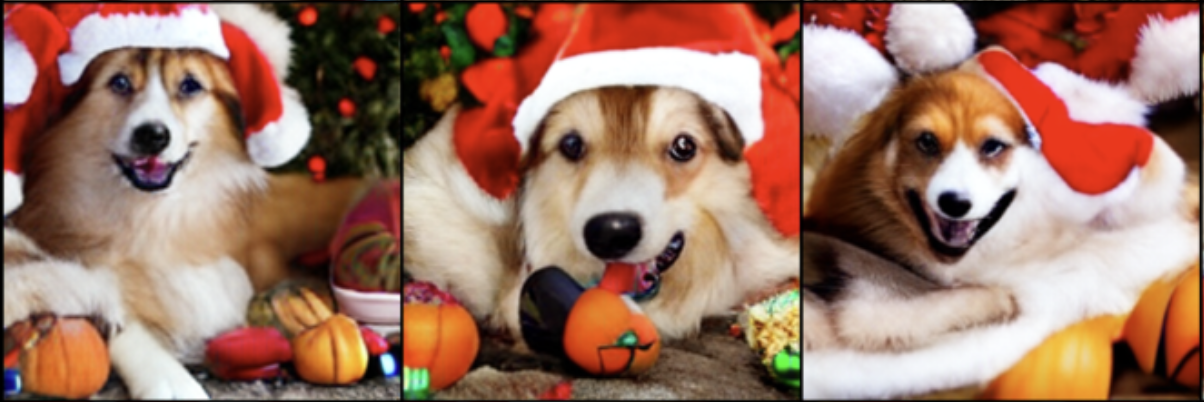}
    \caption{Images generated by adopting the MoT-style~\citep{liang2024mixture} fusion.}
    \label{fig:mot_ex}
    \vspace{-5mm}
\end{wrapfigure}

Mixture-of-Transformers (MoT)~\citep{liang2024mixture} has been widely adopted in unified multimodal models~\citep{shi2024lmfusion,deng2025emerging,liao2025mogao}, including BAGEL~\citep{deng2025emerging}.
When applying MoT to combine a pretrained VLM with TARFlow, we find two failure modes:
(1)~Freezing the VLM and training only the TARFlow-specific branch leads to inferior generation (\cref{fig:mot_ex}), potentially because horizontal MoT-style fusion is ill-suited to single-pass causal autoregressive flows: unlike diffusion models, TARFlow cannot iteratively incorporate VLM conditioning across layers and instead relies mainly on same-layer attention;
(2)~Jointly finetuning the VLM degrades understanding (MME drops to $\sim$800), suggesting that naively adapting VLM parameters for flow-based generation risks erasing pretrained understanding ability before learning effective unified generation.
These observations motivate \pretzel{}: by vertically interleaving a frozen VLM with a trainable TARFlow stream through zero-initialized skip connections, \pretzel{} preserves pretrained understanding while enabling the flow to access rich VLM representations at every position.
Empirically, \pretzel{} improves generation while maintaining substantially stronger understanding than MoT.


\subsection{Analysis of Vertical Skip Connections}

The \pretzel{} architecture employs the vertical skip connections to allow information exchange between the VLM and TARFlow stream.
Since these connections are activated only in Stage~3 with zero-initialized projections, as described in \cref{sec:pipeline}, we therefore examine whether the later-activated connections become effectively used after training.

We first focus on the visual vertical skip connection at the TARFlow input in \cref{eq:vis_res}, which injects the VLM representation $\vy_{\mathrm{vlm},t}$ into TARFlow at each visual position $t \in \mathcal{N}$. 
Specifically, we measure the contribution ratio $r_t^{\mathrm{vis}}$ and directional alignment $s_t^{\mathrm{vis}}$ between the intermediate TARFlow visual latent and the projected VLM feature as follows:
\begin{equation}
    r_t^\mathrm{vis} =
    \frac{
        \|\mW_{\mathrm{vlm}}\vy_{\mathrm{vlm},t}\|_2
    }{
        \|\vu_t\|_2 + \|\mW_{\mathrm{vlm}}\vy_{\mathrm{vlm},t}\|_2
    },
    \quad
    s_t^\mathrm{vis} = \cos\!\left(\vu_t, \mW_{\mathrm{vlm}}\vy_{\mathrm{vlm},t}\right), \quad t \in \mathcal{N}.
\end{equation}

\begin{figure}[!t]
    \centering
    \includegraphics[width=\linewidth]{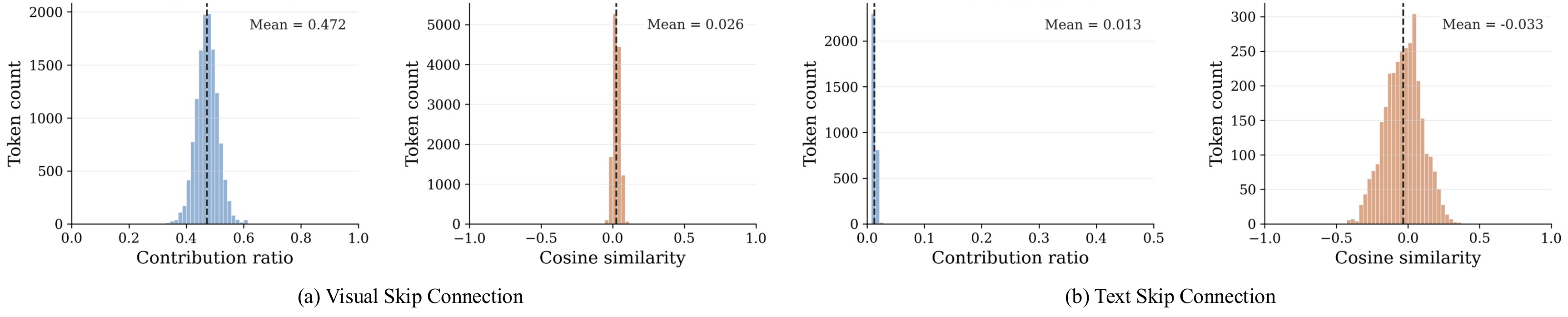}
    \vspace{-2mm}
    \caption{\textbf{Analysis of the vertical skip connection.} 
    }
    \label{fig:two_stream}
\end{figure}

We perform text-to-images generation from 50 randomly sampled text prompts and visualize the distributions of these two quantities in \cref{fig:two_stream} (a). 
We observe that the contribution ratio has a mean of $0.472$, indicating that the projected VLM stream accounts for a substantial fraction of the fused representation magnitude. 
Meanwhile, the near-zero cosine similarity suggests that the VLM stream contributes complementary information through the vertical skip connection.

Similarly, for the textual vertical skip connection at the output in \cref{eq:txt_res}, we analyze how much the TARFlow stream contributes to the final language-modeling representation. 
For each text position $t \in \mathcal{M}$, we define the contribution ratio $r_t^{\mathrm{txt}}$ and the cosine similarity $s_t^{\mathrm{txt}}$:
\begin{equation}
    r_t^{\mathrm{txt}} =
    \frac{
        \|\mW_{\mathcal{D}}\vy_{\mathcal{D},t}\|_2
    }{
        \|\vy_{\mathrm{vlm},t}\|_2 + \|\mW_{\mathcal{D}}\vy_{\mathcal{D},t}\|_2
    },
    \quad
    s_t^{\mathrm{txt}} =
    \cos\!\left(\vy_{\mathrm{vlm},t}, \mW_{\mathcal{D}}\vy_{\mathcal{D},t}\right), \quad  t \in \mathcal{M}.
\end{equation}
As shown in \cref{fig:two_stream} (b), the textual skip connection exhibits a much smaller contribution ratio, suggesting that the projected TARFlow output states $\mW_{\mathcal{D}}\vy_{\mathcal{D},t}$ only lightly corrects the pretrained VLM representation. 
This is consistent with the design goal of preserving the pretrained language modeling capability while allowing TARFlow to provide modest multimodal corrections.

\section{Related Work}

\paragraph{Generative Modeling Paradigms.}
Text generation is dominated by autoregressive LLMs~\citep{achiam2023gpt}, while visual generation is led by diffusion and flow-matching methods~\citep{ho2020denoising,rombach2022high,peebles2023scalable,lipman2023flow,esser2024scaling} whose iterative sampling is structurally distinct from single-pass autoregressive decoding.
Discrete tokenization~\citep{van2017neural,yu2024language,luo2024open} bridges this gap but introduces quantization loss.
Normalizing flows~\citep{dinh2014nice,rezende2015variational,kingma2018glow,ho2019flow++} offer exact likelihood and single-pass sampling; recent TARFlow-style models~\citep{zhainormalizing,gu2025starflow,gu2025starflowv} parameterize flows with causal Transformers, matching diffusion quality while sharing the same left-to-right structure as LLMs.
\methodname{} extends autoregressive flows from vision-only generation to unified multimodal modeling for the first time.

\vspace{-1em}

\paragraph{Unified Multimodal Models.}
A prominent approach combines autoregressive language modeling with diffusion for images~\citep{zhou2024transfusion,xie2024show,xie2025show,shi2024lmfusion,liu2025tuna,liu2026tuna,deng2025emerging,liao2025mogao}, but inherits a structural asymmetry: text tokens enter the KV-cache causally while images require iterative denoising and re-encoding for interleaved generation.
MoT~\citep{liang2024mixture}, adopted in BAGEL~\citep{deng2025emerging}, routes modalities to separate feed-forward parameters---a horizontal separation that maintains two sub-networks within one shell.
Discrete unified approaches~\citep{wang2024emu3,li2025onecat,chen2025janus,chen2025blip3} avoid the hybrid design but sacrifice continuous fidelity.
\methodname{} achieves true unification via the \pretzel{} architecture, which vertically interleaves TARFlow and VLM streams under the same causal mask with skip connections, avoiding both re-encoding overhead and routing complexity.
\vspace{-1em}

\paragraph{Visual Representations.}
Many unified models decouple understanding and generation representations~\citep{chen2025janus,xie2024show,tong2024metamorph}, while recent work explores shared representations~\citep{liu2025tuna,qu2025tokenflow}.
\methodname{} operates in the FAE latent space~\citep{gao2025one}, which provides compact continuous latents serving both understanding and flow-based generation within a single representation.

\section{Conclusion}

We presented \methodname{}, a unified multimodal model that bridges language models and normalizing flows under the same causal Transformer mechanism via the \pretzel{} architecture.
By vertically interleaving a frozen pretrained VLM with a TARFlow stream through residual skip connections, \methodname{} simultaneously satisfies three desiderata: preserving pretrained multimodal understanding (D1), generating high-fidelity continuous images without quantization (D2), and unifying both modalities under a single causal mechanism without diffusion's iterative denoising (D3).
Together with a deep--shallow TARFlow design and a unified FAE latent space, the architecture supports multimodal understanding, text-to-image generation, image editing, and interleaved text--image generation with cache-friendly inference.
Experiments show that \methodname{} achieves strong image generation (0.82 GenEval, 84.14 DPG-Bench) while retaining the pretrained VLM's multimodal capabilities---and that joint training further improves generation by 60.8\% on GenEval relative to the generation-only stage.
These results establish autoregressive normalizing flows as a principled foundation for unified multimodal modeling.
Scaling to higher resolutions, end-to-end training across all components, and improving fine-grained visual fidelity remain important directions for future work (see \cref{sec:limitation}).

\newpage

\bibliographystyle{plainnat}
\bibliography{main}

\appendix
\section{Limitations and Future Work}
\label{sec:limitation}

While \methodname{} demonstrates the potential of TARFlow-style normalizing flows for unified multimodal modeling, several limitations remain. 
First, \methodname{} relies on a multi-stage training pipeline to stably integrate the pretrained VLM, FAE visual representation, adapter, and TARFlow components. Although effective, this staged procedure introduces additional complexity and may lead to under-optimization. A natural direction for future work is to optimize all components end-to-end, allowing the visual representation and cross-modal fusion modules to be jointly shaped by both next-token prediction and TARFlow-based likelihood objectives.

Second, the current model is constrained by the pretrained FAE encoder. In particular, the image resolution and visual quality are limited by the FAE latent space, which can affect fine-grained visual fidelity and text rendering. Replacing the pretrained FAE encoder with a more native visual representation, such as pixel-level or patch-level embeddings, is a promising direction. This would reduce dependence on an external visual tokenizer or autoencoder and move \methodname{} toward a more fully native unified multimodal model.

Finally, although \methodname{} supports multimodal understanding, image generation, editing, and interleaved text--image generation in a single causal framework, it is not yet state-of-the-art on all benchmarks. Improving data scale, training stability, visual representation learning, and long-context interleaved generation remains important future work.
Nevertheless, our results suggest that autoregressive normalizing flows offer a promising foundation for unified multimodal modeling, providing a new direction that combines continuous visual generation, exact likelihood training, and cache-friendly causal decoding within a single architecture.

\section{Impact Statement}
The proposed method explores autoregressive normalizing flows as a foundation for unified multimodal understanding and generation. By enabling text and visual latents to be generated under the same causal framework, this work may contribute to more efficient and flexible multimodal systems, particularly for interleaved text-image generation, image editing, and interactive visual reasoning. 
More broadly, unified multimodal models could improve accessibility and communication by helping users express ideas across modalities, generate visual explanations, and interact with information in more natural ways. They may also support applications in domains such as media production, data visualization, simulation, and assistive technologies.

\section{Implementation Details}
\label{sec:imp}

\subsection{Architecture Design}
\label{sec:appendix-arch}

\begin{table}[h]
\centering
\small
\setlength{\tabcolsep}{6pt}
\begin{tabular}{ll}
\toprule
Component & Specification \\
\midrule
Pretrained VLM & Qwen2.5-VL-7B-Instruct~\citep{Qwen25-VL} (frozen) \\
FAE  & FAE~\citep{gao2025one} on DINOv2-g/14~\citep{oquab2023dinov2} (frozen) \\
Deep TARFlow $f_{\mathcal{D}}$ &  24 Transformer Layers, width $3072$ \\
Shallow TARFlows $f_{\mathcal{S}}$ & 2 Blocks, 4 Transformer Layers each Block, width $3072$ \\
Visual adapter & 1 MLP, 1 FiLM Layer \\
Trainable parameters & 3.6B \\
\bottomrule
\end{tabular}
\vspace{1em}
\caption{
Model specification of \methodname{}. The pretrained VLM and FAE encoder are kept frozen.
}
\label{tab:arch}
\end{table}

\subsection{Training Details}

\methodname{} is trained on $64$ H100 GPUs. In all the experiments, we share the following training configuration for our proposed \methodname{}. 
\begin{verbatim}
training config:
    batch_size=1024
    optimizer='AdamW'
    adam_beta1=0.9
    adam_beta2=0.95
    adam_eps=1e-8
    min_learning_rate=1e-6
    learning_rate_schedule=cosine
    weight_decay=1e-4
    mixed_precision_training=bf16
\end{verbatim}
We use a learning rate of $1e-4$ for Stage 1 and Stage 2 training and a learning rate of $5e-5$ for Stage 3

\section{Qualitative Examples}

\begin{figure*}[!t]
    \centering
    \includegraphics[width=0.9\linewidth]{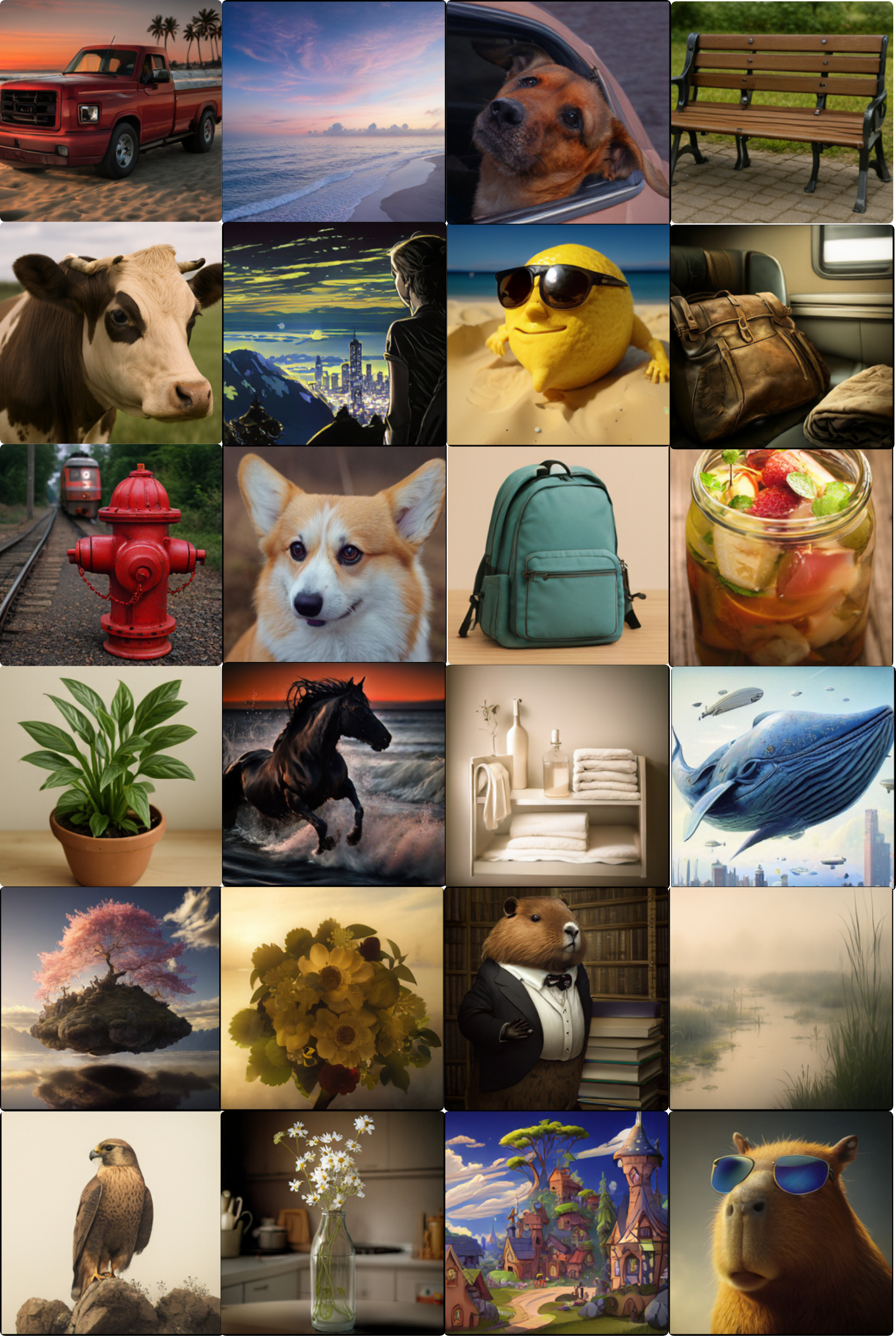}
    \vspace{-4mm}
    \caption{Text-to-Image generation examples from \methodname{} at 256 $\times$ 256 resolution.}
    \label{fig:t2i_ex_more}
    \vspace{-3mm}
\end{figure*}
\begin{figure*}[!t]
    \centering
    \includegraphics[width=0.9\linewidth]{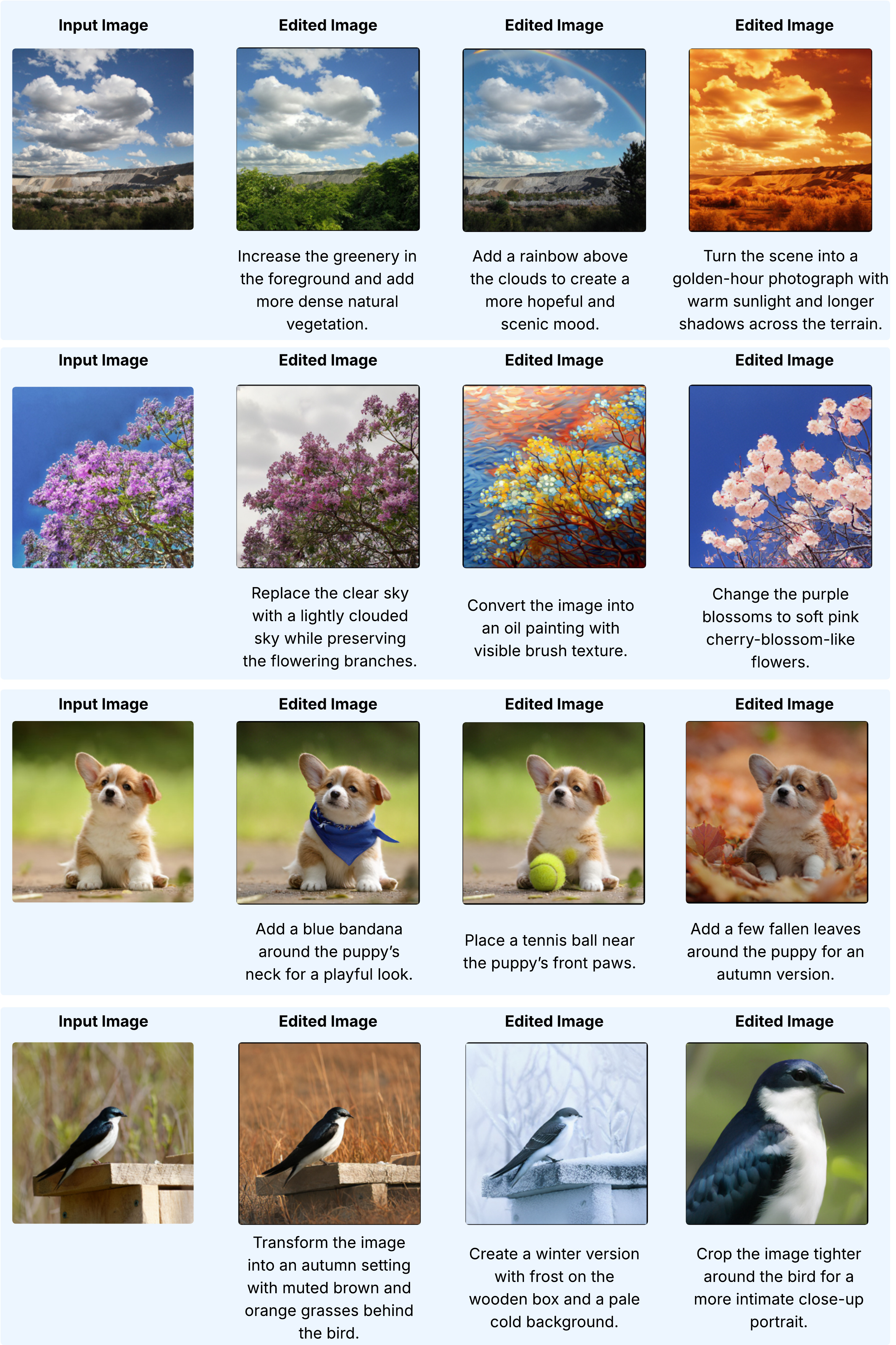}
    \vspace{-4mm}
    \caption{Image editing examples from \methodname{}.}
    \label{fig:edit_more}
    \vspace{-3mm}
\end{figure*}

\cref{fig:t2i_ex_more,fig:edit_more} shows qualitative examples for text-to-image generation and image editing examples.

\applefootnote{ \textcolor{textgray}{\sffamily Apple and the Apple logo are trademarks of Apple Inc., registered in the U.S. and other countries and regions.}}

\end{document}